\pdfoutput=1

\documentclass[11pt]{article}

\usepackage[]{acl}

\usepackage{times}
\usepackage{latexsym}
\usepackage{graphicx}
\usepackage{amsmath}
\usepackage{multirow}
\usepackage{array}

\usepackage[T1]{fontenc}

\usepackage[utf8]{inputenc}

\usepackage{microtype}

\usepackage{inconsolata}

\title{Gazetteer-Enhanced Bangla Named Entity Recognition with BanglaBERT Semantic Embeddings K-Means-Infused CRF Model}

\author{\begin{normalsize}Niloy Farhan$^*$, Saman Sarker Joy$^*$, Tafseer Binte Mannan, Farig Sadeque\end{normalsize} 
\\ \begin{normalsize}Department of Computer Science and Engineering\end{normalsize} \\
\begin{normalsize}BRAC University \end{normalsize} \\
\begin{normalsize}Merul Badda, Dhaka - 1212, Bangladesh\end{normalsize}\\
\begin{small}\texttt{\{niloy.farhan, saman.sarker.joy, tafseer.binte.mannan\}@g.bracu.ac.bd, farig.sadeque@bracu.ac.bd}\end{small}
}

\newcommand\blfootnote[1]{%
  \begingroup
  \renewcommand\thefootnote{}\footnote{#1}%
  \addtocounter{footnote}{-1}%
  \endgroup
}

\begin{document}
\maketitle
\begin{abstract}
Named Entity Recognition (NER) is a sub-task of Natural Language Processing (NLP) that distinguishes entities from unorganized text into predefined categorization. In recent years, a lot of Bangla NLP subtasks have received quite a lot of attention; but Named Entity Recognition in Bangla still lags behind. In this research, we explored the existing state of research in Bangla Named Entity Recognition. We tried to figure out the limitations that current techniques and datasets face, and we would like to address these limitations in our research. Additionally, We developed a Gazetteer that has the ability to significantly boost the performance of NER. We also proposed a new NER solution by taking advantage of state-of-the-art NLP tools that outperform conventional techniques.
\end{abstract}

\section{Introduction}
\begin{figure}[]
\centering
\includegraphics[width=0.5\textwidth]{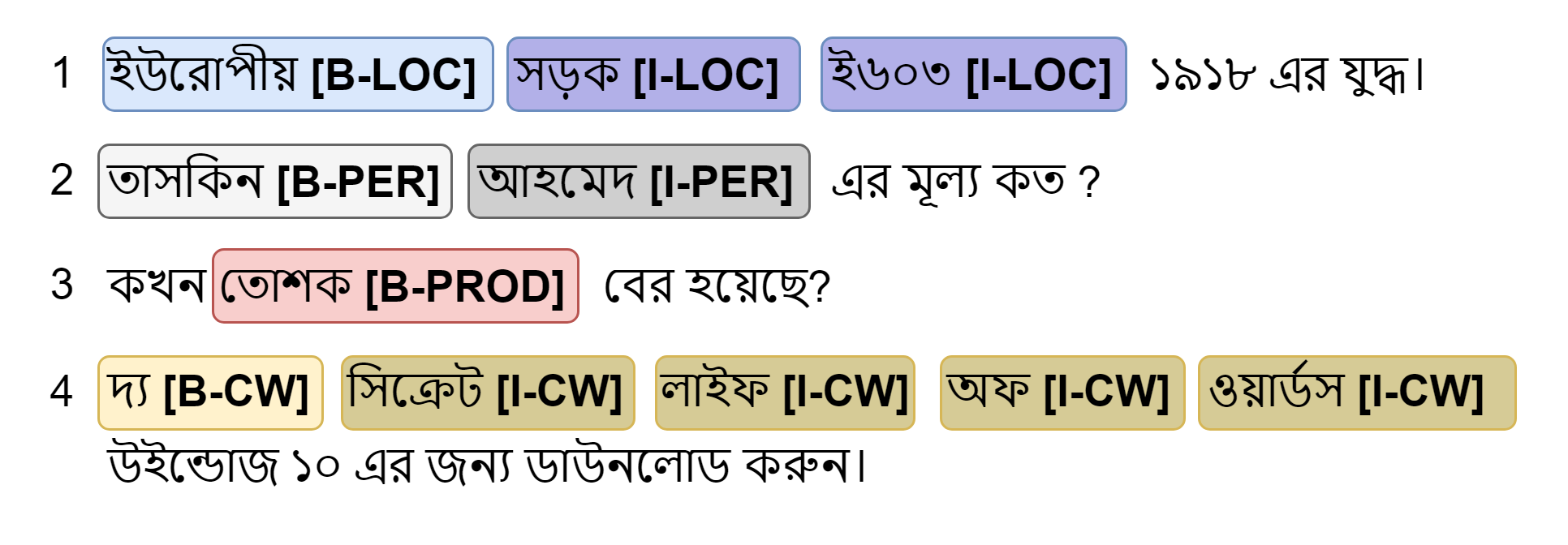}
\caption{NER Example in Bangla}
\label{fig:bangla_ner}
\end{figure}

\blfootnote{$^*$ Equal Contributions}
Named Entity Recognition (NER) is a significant initial step in Natural Language Processing (NLP) tasks. It has been widely used in machine translation, information extraction, automatic question-answering, natural language understanding, text-to-speech synthesis, etc. An example of Bangla NER is shown in figure \ref{fig:bangla_ner}.

NER systems can use a Rule-based approach, Machine Learning (ML) or Deep Learning (DL) approach, or using Hybrid approach. These are the only guidelines that a rule-based NLP system needs to classify the language and to analyze it \cite{1}. This method can be used for a particular language depending on its grammatical structure. Compared to machine learning techniques, this type of NE (Named Entitiy) Rule Base approach exhibits greater accuracy. But this is hard to maintain as it is language dependent and the set of rules can be huge and complicated. Furthermore, the use of ML and DL is more popular in NER systems as these are easy to train and less expensive. A hybrid NER system is made by merging the rule-based method and the statistical ML or DL methods.

Despite being the seventh most popular language \cite{2} very few notable works have been done in the field of NLP in Bangla. We examined the current state of Named Entity Recognition research in Bangla in this research. To address these limits in our research, we had investigated the limitations that current approaches and datasets confront. In our research, we had discussed the poor performance of typical deep-learning methods and came up with the idea of a knowledge-based approach.  We created Gazetteer which is a set of entity lists of each tag that can be used as an external knowledge source for neural network models which we made publicly available. Our Gazetteer showed promising improvement in the NER classification task. Furthermore, we have experimented with BanglaBERT models and tried to improve them with our custom weighted cross entropy loss function which we have discussed in detail in the upcoming section. Finally, we experimented with different models with different features and strategies and based on our findings, we have proposed a completely new NER solution which is a CRF model with Gazetteer and BanglaBERT embeddings significantly outperforming conventional techniques.
All the necessary codes, gazetteers data, models and other experiments are available at \url{https://github.com/samanjoy2/gazz-ban-ner}.\\\\
The main contributions of this paper can be summarized as follows:
\begin{itemize}
\item We did extensive Exploratory Data Analysis on a state-of-the-art Bangla NER dataset; MultiCoNER I
\item We developed a Bangla gazetteer with 93,749 data that can be used in any deep learning framework for providing a knowledge-based solution
\item We experimented with BanglaBERT models and tried to improve them with our custom weighted cross entropy loss function 
\item We proposed a new NER solution; a CRF model with Gazetteer and BanglaBERT embeddings that significantly outperforms conventional techniques
\end{itemize}

\section{Related Works}
Over many years lots of research work has been done with NER in different languages. Recently, several supervised learning-based models and rule-based models were proposed.

In the paper \cite{3}, the authors experimented with the combination of three approaches: Dictionary-based NE detection, Rule-based NE detection, and n-gram based NE detection. They created the dataset from Bangla daily newspaper. Their model was able to achieve an 89.51\% $F_{1}$ score.

In another study \cite{4}, the author proposed a machine learning-based model for the NER system. Their corpus had 10,000 words that were taken from 690 sentences in total. They received an $F_{1}$ score of 68.98\% from the MEMM model alone and an $F_{1}$ score of 71.59\% from the combined model.

In their research \cite{6}, the authors focused on POS tagging and Name Entity Recognition in the Bangla language. They created two distinct datasets for POS tagging and NER by compiling news articles from different Bangla online news portals. For this work, deep neural network models have been implemented - Bi-directional Long short-term memory (BLSTM), Convolutional Neural Network (CNN), Conditional Random Field (CRF), and LSTM. They achieved a 0.6285 $F_{1}$ score with their model.

According to the author of the \cite{7} paper, they experimented with eleven different architectures consisting of BERT (Bidirectional Encoder Representations from Transformers) layer, BiLSTM layers, linear layers, a dropout layer in between and a CRF layer. They used the dataset from Karim et al. \cite{8} where 72000 Bangla sentences were annotated. Their highest score was achieved with BERT + BiLSTM + CRF + CW with a macro averaged $F_{1}$ score of 65.96\%.

There are some notable works done in order to build NER annotated corpus.

The \cite{9} paper is dedicated to a corpus annotated with seven named entities (Person, Location, Organization, Facility, Time, Units, Misc.) with a discreet attention on Bangladeshi Bangla. They have worked on baseline results as well. They collected newspaper stories from several newspapers and annotated them for the dataset.

In the \cite{10} research paper, they worked with a large curated Bangla Articles Dataset and many supervised learning models were proposed for analyzing these data. According to them, there is a very a smaller number of curated Bangla datasets so they themselves curated a huge dataset from Bangla articles from different news sites containing 3,76,226 articles. Applying to pre-process (removal of punctuations, stopwords, etc.) they got a set of words frequency. TF-IDF and Word2Vec approach has been used in this paper. Although this is a large dataset, it is not annotated for the NER task.

From the above discussion, it is observed that in the above research papers mostly deep learning models like Long Short-Term Memory (LSTM), and Bi-directional Long short-term memory (BLSTM) have been used for Bangla name entity recognition. And not every model could give perform reliably enough performance like English, Chinese or other languages. As Bangla NER is a very unique work, the technique should not be limited. Recently transformer models of deep learning have come to light, such new technology should be implemented for a better and more accurate score.

In the \cite{wang2022damonlp} research paper, the author proposed a knowledge based system. They were working with the MultiCoNER 1 dataset which is similar to us. This approach wins 10 out of 13 categories in the MultiCoNER task. They were able to achieve 0.8351 Macro $F_{1}$ score in the test data.

The \cite{bhattacharjee-etal-2022-banglabert} paper introduced us with BanglaBERT and BanglishBERT. BanglaBERT has a base and a large model both of which are based on BERT. They pretrained BanglaBERT using ELECTRA. For zero-shot cross-lingual transfer, their BanglishBERT achieved 55.56 Micro $F_{1}$ score and for supervised fine-tuning, their BanglaBERT model gained 77.78\% Micro $F_{1}$ score in MultiCoNER dataset.

In another study \cite{panchendrarajan-amaresan-2018-bidirectional}, the author presented a novel method on Named Entity Recognition (NER) using a Bidirectional Long Short-Term Memory (LSTM) Conditional Random Field (CRF) model. The proposed model’s efficacy is demonstrated in the study through experiments on benchmark datasets, highlighting both its capacity for cutting-edge performance in NER tasks. Their model was able to achieve a 90.84\% $F_{1}$ score.
\section{Dataset}
\label{sec:dataset}
\subsection{MultiCoNER I Dataset}
\label{subsec:dataset}
\begin{figure}[h]
\centering
\includegraphics[width=0.5\textwidth]{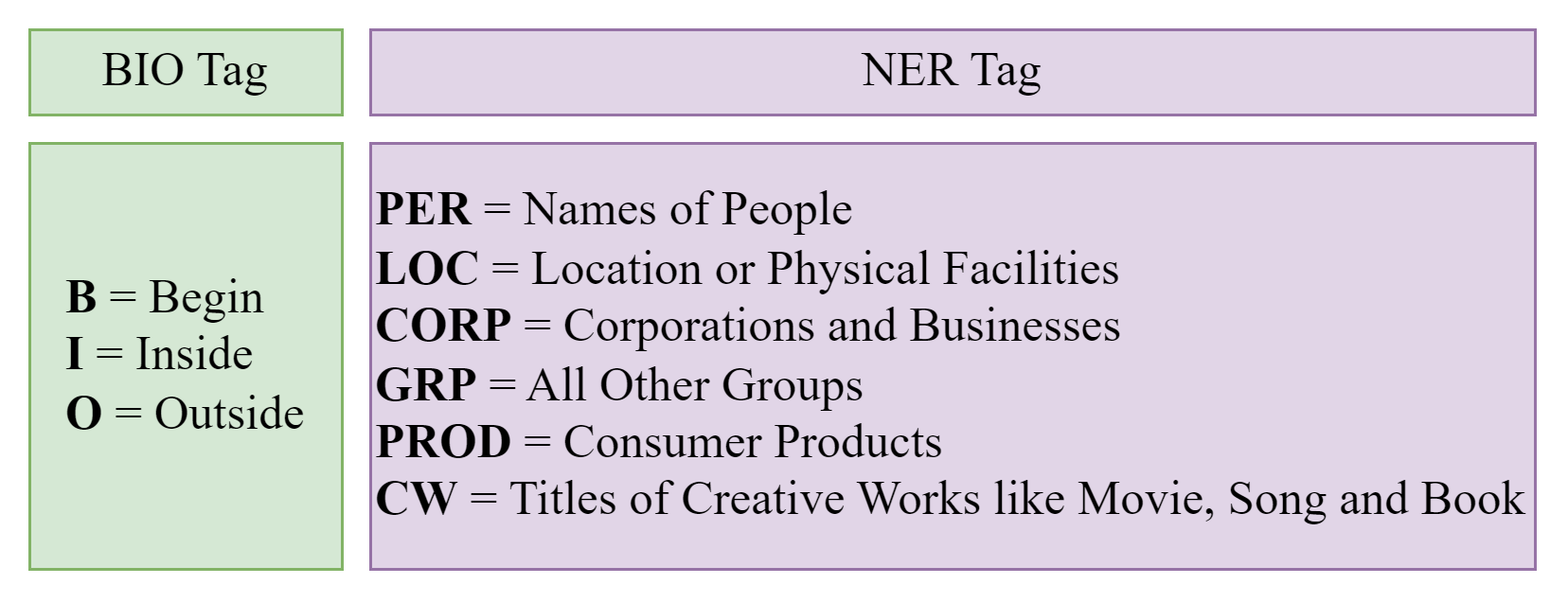}
\caption{BIO Tags}
\label{fig:biotagl}
\end{figure}

The dataset we decided to use is MultiCoNER I \cite{MultiCoNER}. The MultiCoNER I dataset we utilized, is a comprehensive multilingual dataset for Named Entity Recognition. MultiCoNER I was built on the WNUT \cite{wnut} entity type taxonomy. The tags name and BIO tags are explained in figure \ref{fig:biotagl}.
\\
Our work focuses solely on the Bangla language. We aimed to work with Bangla NER. The class distribution of train, test and dev of Bangla data is presented on table \ref{tab:class}.


\begin{table}[h]
\centering
\begin{tabular}{lccccccc}
\hline
\textbf{Class} & \textbf{Train} & \textbf{Dev} & \textbf{Test} \\
\hline
\textbf{PER} & 2,606 & 144 & 24,601 \\
\textbf{LOC} & 2,351 & 101 & 29,628 \\
\textbf{GRP} & 2,405 & 118 & 19,177 \\
\textbf{CORP} & 2,598 & 127 & 20,066 \\
\textbf{CW} & 2,157 & 120 & 21,280 \\
\textbf{PROD} & 3,188 & 190 & 20,878 \\
\hline
\textbf{Total} & 15,300 & 800 & 133,119 \\
\hline
\end{tabular}
\caption{Class Distribution of Train Test and Dev of Bangla Data}
\label{tab:class}
\end{table}

\subsection{Gazetteers} \label{gz}

\begin{figure}[h!]
\centering
\includegraphics[width = 0.5\textwidth]{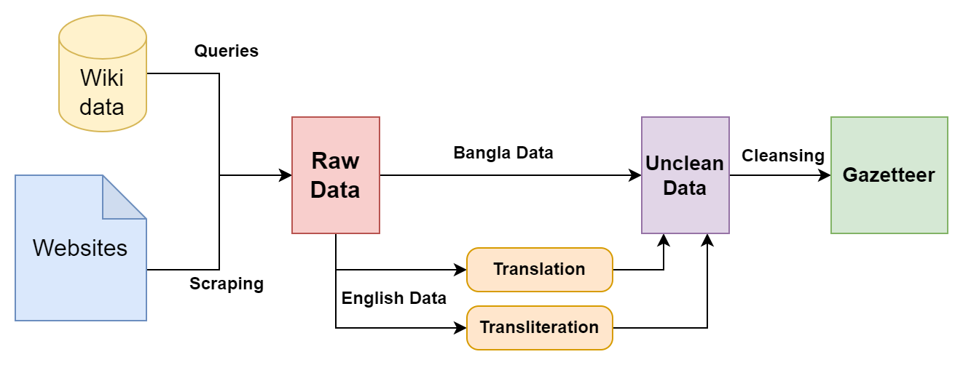}
\caption{Workflow of the formation of our gazetteer}
\label{fig:gazzformationall}
\end{figure}

The knowledge-based solution that we mentioned earlier is Gazetteer. The word “gazetteer” is used interchangeably for both the set of entity lists and for the processing resource that uses those lists to find occurrences of named entities in texts. There has been much research on this Gazetteer technique and it has shown promising improvements to tasks like NER. From the paper \cite{Chen_2022}, we can see that the implementation of gazetteer has an impact on the overall performance of their NER system. Building Gazetteer for Bangla is a difficult task as the resource are not rich in Bangla. We have created a Gazetteer for Bangla in this research. The whole workflow of the formation of our gazetteer is in figure \ref{fig:gazzformationall}.

\subsubsection{Extracting Bangla Data from Wikidata and Scraping}
At first, we have collected bangla data for each of the tags (LOC, PER, CORP, GRP, PROD, CW) from Wikidata\footnote{\url{https://www.wikidata.org/}}. We have used Wikidata Query Service\footnote{\url{https://query.wikidata.org/}} where we have put different queries written in SPARQL language and got our desired data. However, we couldn't get satisfactory amount of data using it for some tags. So, we also scraped a lot of websites including Wikipedia and newspaper portals to gather Bangla data. However, as those data were randomly gathered, we had to manually hand-pick the data according to our tags. We especially did this for PROD and CW as the number of Bangla Wikidata of these were significantly low. 
\subsubsection{Converting the English Data}
While collecting data, we found that the amount of English data on this was enormous. So, like how we gathered the Bangla Data, we also gathered a lot of English data according to the tags required. Then we converted the English data to Bangla by following 2 methods:

\textbf{Translation:} We translated the English data to Bangla data. We did this using the help of \textit{Google API} translation tool \footnote{\url{https://cloud.google.com/translate}}.

\textbf{Transliteration:} Transliteration refers to the method of mapping from one system of writing to another based on phonetic similarity. We transliterated the English data to Bangla data as some English word is used in Bangla by same pronunciation. 
\subsubsection{Cleaning the data}
As we were working on a large list of words, there were some broken words and some unnecessary punctuations here and there. We cleaned the data by removing all these unnecessary punctuations and broken words. Then we also removed any duplicate words.

After collecting all $93,749$ data, the distribution of each tags is given in table \ref{tab:tagsingazz}.

\begin{table}[h]
\centering
\begin{tabular}{lc}
\hline
\textbf{NER Tags} & \textbf{Number of Data} \\
\hline
PER & 26296 \\
CW & 20446 \\
LOC & 16617 \\
CORP & 13737 \\
GRP & 10517 \\
PROD & 6136 \\
\hline
\end{tabular}
\caption{Total Number of Each Tag in our Gazetteer}
\label{tab:tagsingazz}
\end{table}

\subsubsection{Trie Data Structure}
This data structure follows a treelike structure where every node represents a single character. This data structure provides the fastest search operations. It is very good at retrieving the strings that include a specific prefix. We created a Trie Data Structure where we inserted all the gazetteer words. It is a treelike data structure that provides fastest search operation. We modified the node class so that it can return us their corresponding NER tags. As we have 93,749 entities, by using Trie DS, we can quickly find if given token is present in gazetteer or not.

We have also implemented a function that takes a Bangla sentence as input and it gives corresponding one-hot vector representation for each word in the sentence which can be used as input in any deep learning framework like PyTorch and TensorFlow.

\begin{figure}[h]
\centering
\includegraphics[width=0.4\textwidth]{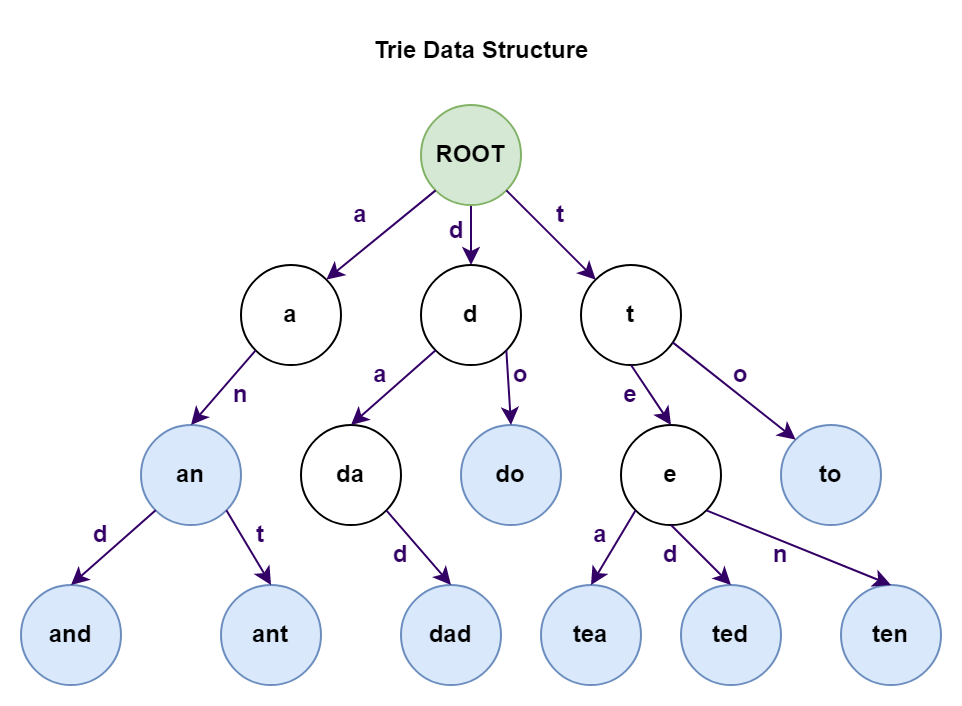}
\caption{Trie Data Structure}
\label{fig:crfmodelfimafdgde}
\end{figure}

\section{Exploratory Data Analysis}
We have done our EDA to MultiCoNER 1 dataset to find its limitations.
\subsection{Large Test Data}
The amount of test data was overwhelmingly larger than the train data, which is mainly done in order to make the model ready to deal with real-world data. However, this means the features from our train data alone are not sufficient for our models to perform a good result which can be reflected in the baseline score (XLM-RoBERTa) of MultiCoNER’s paper \cite{MultiCoNER}.
\subsection{Irregularities in the Punctuations}
We noticed that in the tokens, there were a lot of irregularities in the punctuations which sometimes caused issues. So, we removed the punctuation from those tokens where they are added as prefixes. We were careful while removing it because some tokens had only one punctuation and were tagged as ‘O’. We did not remove those punctuation which were a single token. We also didn’t remove the period in abbreviations of the words.
\subsection{Presence of Foreign Words}
In the dataset, we also found words of some other languages such as Hindi, Farsi, English, Chinese, and Russian. The presence of foreign language data in Bangla Dataset can be caused by translation errors as we have discussed in the formation of the MultiCoNER I dataset.



\subsection{Imbalanced Dataset}
One common problem of NER is,  the most frequent class is ‘O’. This causes an imbalance in the dataset which makes model bias towards majority classes. In the MultiCoNER dataset, this problem was more extreme because 83.5\% the ‘O’ tag in training data.
\subsection{Alignment of tags after tokenization}
Models like BanglaBERT base and BanglaBERT large use their BanglaBERT tokenizer. This tokenizer takes a sentence and breaks it into tokens with their input ids and attention masks. However, this created a problem of misalignment of the tags in the sentence which can create error in loss and metric function. To fix this issue, we aligned our tags corresponding to their tokens.

\section{Models}
We divided this section into three parts. Baseline models showed the current state of deep learning models in Bangla NER classification. Secondly, we talked about BanglaBERT which is currently state-of-the-art BERT model for Bangla and a potential solution to improve them. Lastly, we talked about proposed model that performed state-of-the-art performance in MultiCoNER I dataset.
\subsection{Baseline Models}
\subsubsection{Two Layer BiLSTM Network} 
We used an embedding layer using Bangla GloVe Embeddings \cite{bnlp} that has 300 dimensions for each token. Then we used our first BiLSTM layer followed by a Feed Forward Neural Network (FFNN). After that, the model has the second BiLSTM Layer which is again followed by the second FFNN. Both BiLSTM and FFNN layers have 200 dimensions. Finally, we used the time distribution layer to get our tags.
\subsubsection{Single Transformer-Based Network} 
We used token and position embedding of 128 dimensions for our word token and pos tag of word tokens. Then the model passes it to the transformer block. The output of the transformer block then goes through the average pooling 2d layer. Here, we used a dropout layer to prevent overfitting. Then, It passes through a fully connected layer with 200 nodes which is also a dropout layer. The final layer is our output layer which is another fully connected neural network.
\subsection{BanglaBERT}
\begin{figure}[h]
\centering
\includegraphics[width = 0.5\textwidth]{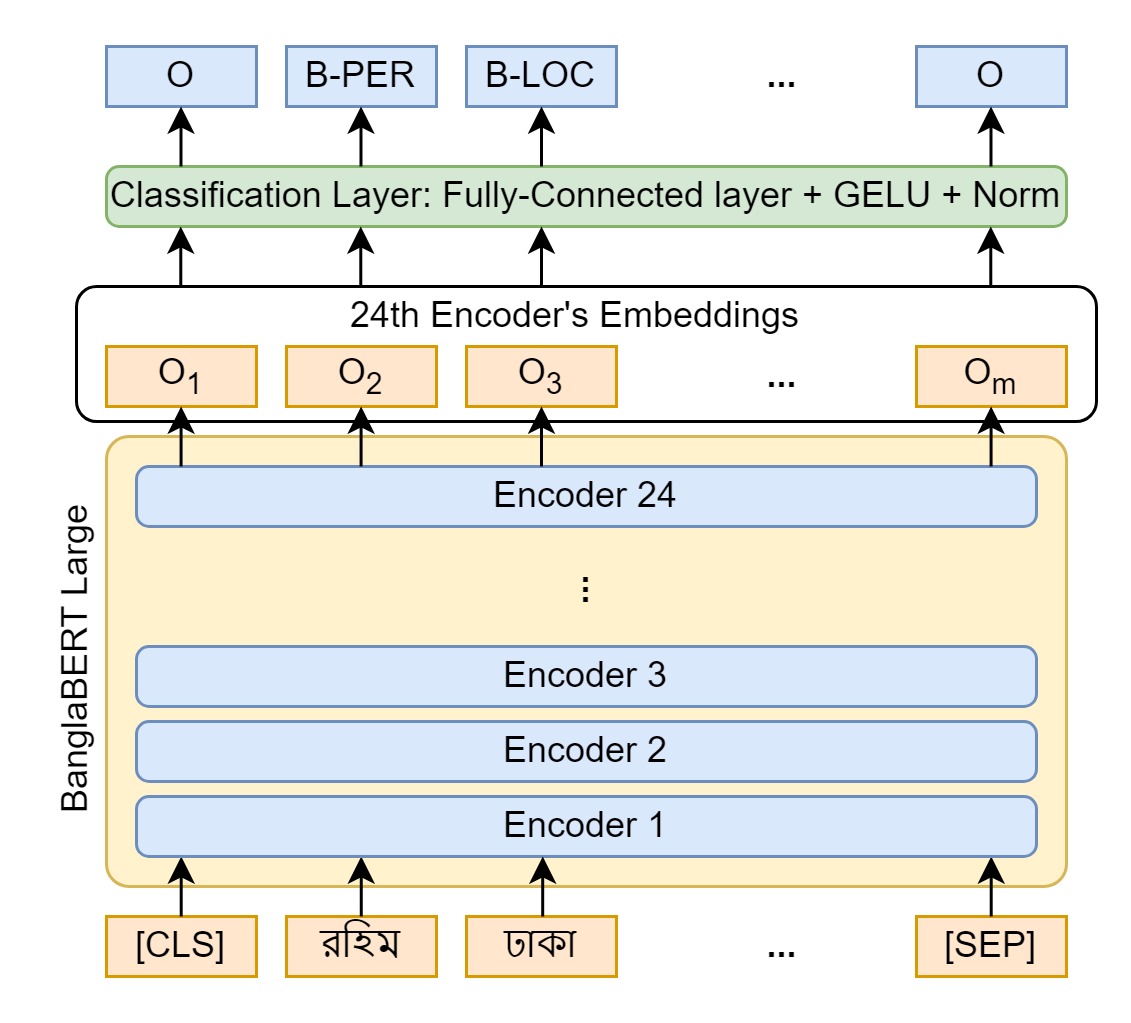}
\caption{Fine-tuned BanglaBERT Large Model}
\label{fig:flargemodel}
\end{figure}

BanglaBERT \cite{bhattacharjee-etal-2022-banglabert} is a pre-trained language model based on ELECTRA.  It is   trained on a large corpus of Bangla text. We used both BanglaBERT Base and BanglaBERT Large in our experiment. The figure \ref{fig:flargemodel} shows our finetuned BanglaBERT large model.
We fine-tuned the pre-trained BanglaBERT large model on the MultiCoNER dataset to perform Named Entity Recognition (NER) in Bangla. Like the base model, we used BanglaBERT tokenizer to get the input ids and attention masks. Then the data was padded using the \textit{DataCollatorForTokenClassification} function where the max sequence length was 64. Similar to the base model, the learning rate was set to 2e-5 and the batch size was 32. We used Pytorch’s Cross-Entropy loss function. At first, we did not use any custom weights for each tag. Later, we used our custom weights for the categorical cross-entropy loss function, which can be found in section \ref{celf}. We used 6 epochs for fine-tuning the model as the score was not improving after that.
\subsubsection{Custom Weighted Categorical Cross Entropy Loss Function} \label{celf}
In attempt to improve current state-of-the-art model, we propose a custom weighted categorical cross entropy loss function. As the dataset was imbalanced, we developed custom loss weights to ensure the model reduces the bias toward more encountering tags. The custom loss weights were calculated using the following formula:
\begin{equation*}(1-\frac{(n + 2) * count(Tag)}{total}) * 10\end{equation*}
Here, n represents number of class, count(Tag) represents the number of that tag in the dataset and the total represnts the total number of tokens in the dataset. The added 2 was for normalizing the weights. 

\subsection{Conditional Random Fields}
CRF is a probabilistic graphical model dedicated to tasks like Named Entity Recognition, part-of-speech tagging, semantic role labeling, and handwriting recognition. It is trained using labeled data, where parameters are learned to maximize the likelihood of observed label sequences. During inference, the most likely sequence of labels is computed based on the learned parameters and observed features. Selection of features is extremely crucial for CRF model.
\begin{figure}[h]
\centering
\includegraphics[width=0.5\textwidth]{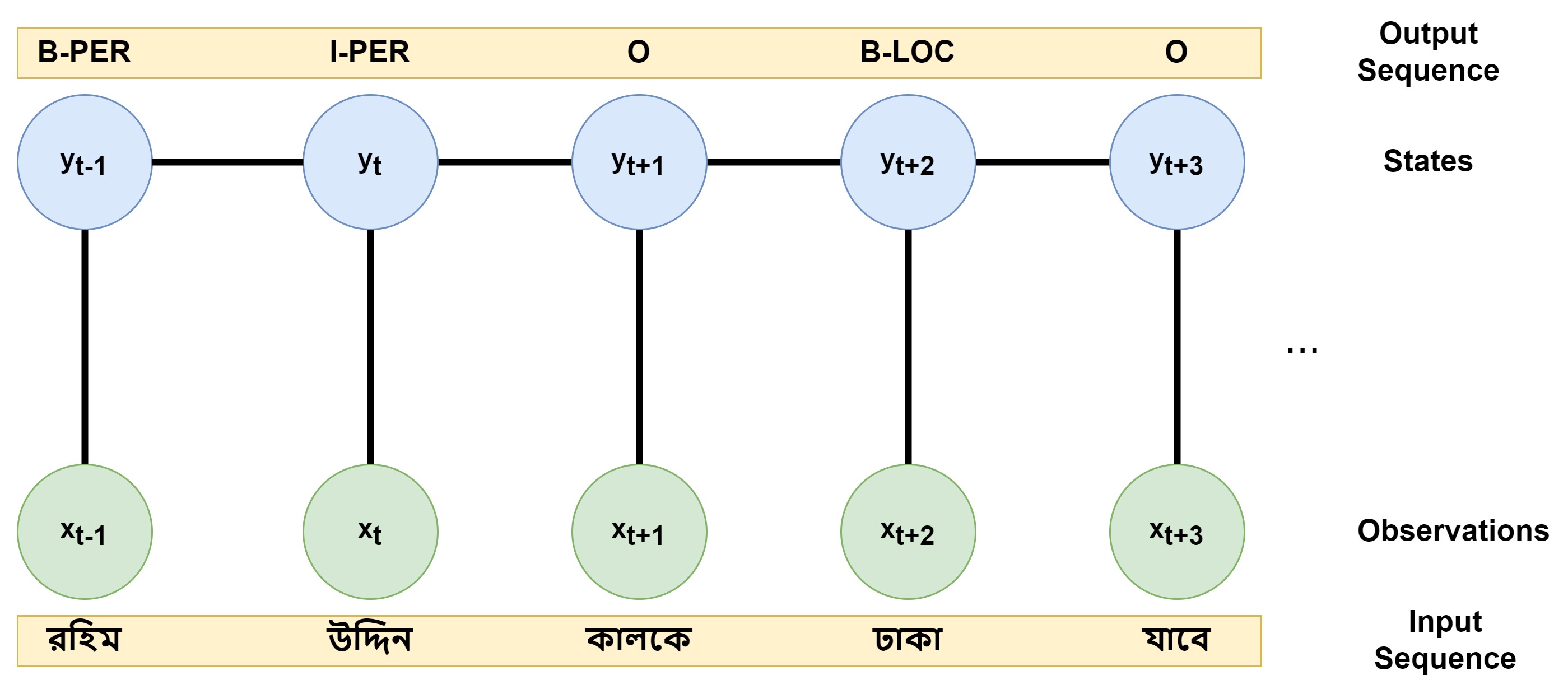}
\caption{CRF}
\label{fig:crf}
\end{figure}
\subsection{Feature Engineering for CRF} 
We experimented with several features for incorporating it with CRF.
\subsubsection{Basic Features}
\textbf{Suffixes}: Last three characters of the word.\\
\textbf{Prefixes:} First three characters of the word.\\
\textbf{Length:} Represents number of character in each word.\\
\textbf{Neighboring Words:} Previous two and next two words.\\
\textbf{IsDigit:} If the token is digit.\\
\textbf{IsPunctuation:} If the word is entirely punctuation (such as commas, periods, question/exclamation marks, etc).\\
\textbf{IsBangla:} If all the character in the token have Unicode between 2432 to 2559 or 32 (Unicode of space) then it is considered as a Bangla word, else it is not a Bangla word.\\
\textbf{IsStopword:} If a word is a stopword, it contains little to no value to the NER model hence it should be excused or should be tagged as 'O'.\\
\textbf{Word Frequency:} Number of occurance of that word in the paragraph or dataset.\\
\textbf{BOS and EOS}: Beginning of sentence (BOS) and End of sentence (EOS).\\
\subsubsection{Part of Speech (POS) Tags}
\begin{figure}[h]
\centering
\includegraphics[width=0.5\textwidth]{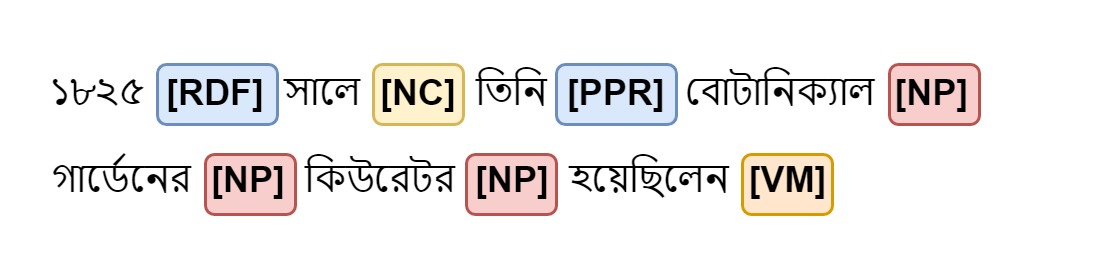}
\caption{Example of POS Tag}
\label{fig:posedddx}
\end{figure}

Incorporating Part of Speech (POS) tags offers valuable insights into the grammatical composition of sentences and the functional roles played by individual words. Example of a POS tag is given in figure \ref{fig:posedddx}.
In our implementation, we leveraged the Bengali POS tag model of BNLP \cite{bnlp}, a CRF-driven POS tagger built upon the NLTR dataset with an 80.75 $F_1$ score. We also added POS tag for neighboring words.
\subsubsection{Gazetteer Information}
We used the idea of Gazetteer Lists (discussed in section \ref{gz}) to further improve our model. As discussed previously in the Dataset section, the Gazetteer we built is a very big and comprehensive list of PER, LOC, GRP, CORP, CW, and PROD. Utilizing gazetteer entries provides explicit labels for otherwise ambiguous text spans, allowing the CRF to make confident judgments when encountering unseen variations. For implementing this feature, we created is\_corp, is\_per, is\_loc, is\_grp, is\_prod, is\_cw for the current word and its neighboring words.
\begin{figure}[h]
\centering
\includegraphics[width=0.5\textwidth]{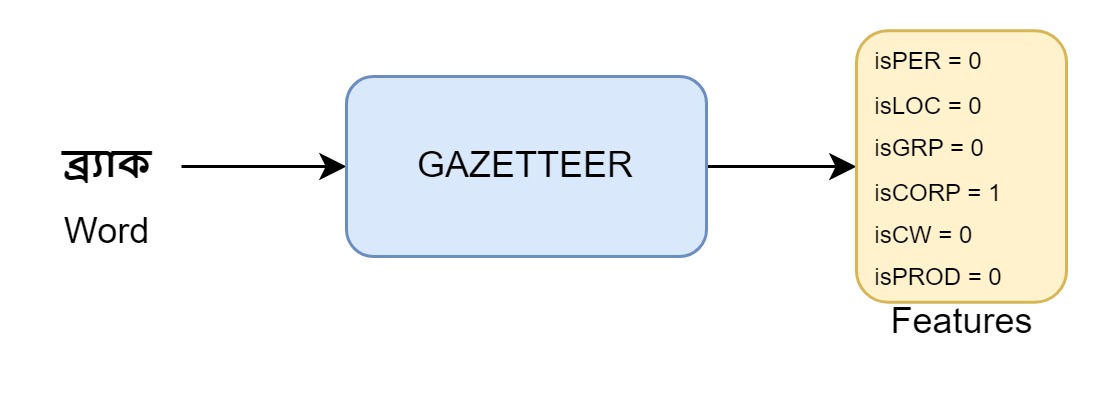}
\caption{Gazetteer as Features}
\label{fig:gazzfeat}
\end{figure}\\
\subsubsection{BanglaBERT Related Features}

\begin{figure}[h]
\centering
\includegraphics[width=0.5\textwidth]{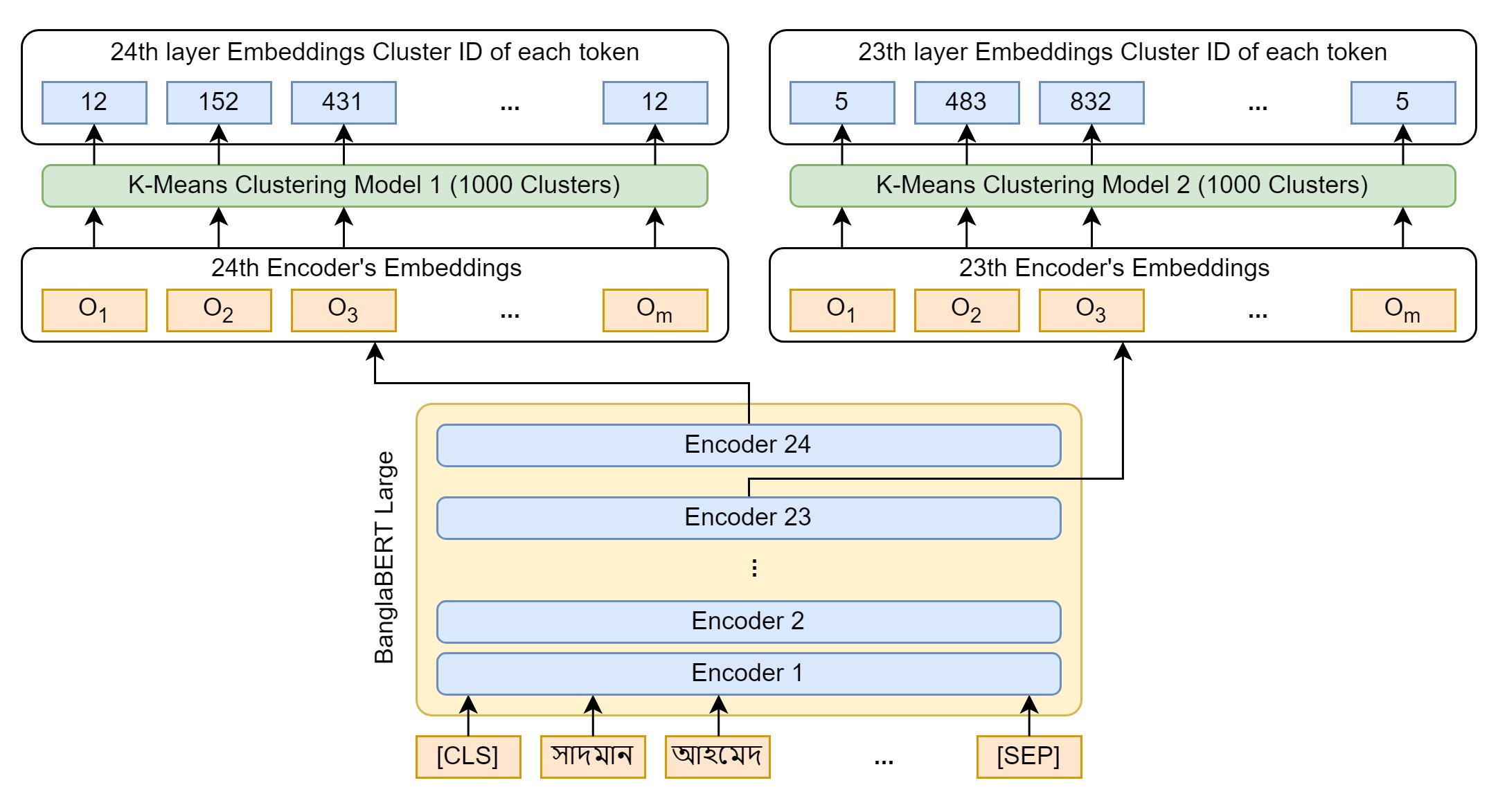}
\caption{BanglaBERT Embeddings with K-means}
\label{fig:bertembkmeans}
\end{figure}

\textbf{BanglaBERT Large Embeddings Cluster Information:} This is a complex feature. At first, we used 23rd and 24th layer of encoders in BanglabBERT large. Then, we used two different kmeans clustering model to with 1000 clusters each. This gives each word two cluster ID for both encoders. A detailed workflow to get this feature is given in \ref{fig:bertembkmeans}.

\textbf{BanglaBERT Large Softmax Outputs:} Here, we pass the sentence to BanglaBERT large model and get its output. Then, the predicted tag by BanglaBERT is added as a feature for each word.

\textbf{Raw BanglaBERT Large 24th Layer’s 1024-Sized Embedding:} We used the raw embedding of BanglaBERT Large as a feature as well. Since the 24th layer encoder produces 1024 sized embedding, we added 1024 features for each word. However, the values of the raw embeddings were mostly in the range of 0 to 1. If we train with these values, then the model will be overfitted with train data, so, to overcome this problem, we transformed the embeddings by using the following formula which eliminates the minor differences.
\begin{equation*}E_{\text {new }}=R_{\text {round }}\left(E_{\text {raw }} * 100\right)\end{equation*}
\subsection{CRF Model Strategies}
We have discussed all the features that we could engineer from the
train data. We did not have a single model out of all these features; 
we experimented with several strategies.\\
\textbf{CRF Model A:} Added Suffix, Prefix, Index, Length. \\
\textbf{CRF Model B:} Added isDigit, isPunctuation, Frequency.\\
\textbf{CRF Model C:} Added POS.\\
\textbf{CRF Model D:} Added Gazetteer.\\
\textbf{CRF Model E:} Added isBangla, isStopword.\\
\textbf{CRF Model F:} Added K-means of BanglaBERTLarge Embeddings of 24th Layer. Removed isBangla, isStopword.\\
\textbf{CRF Model G:} Added K-means of BanglaBERTLarge Embeddings of 23rd and 24th Layer.\\
\textbf{CRF Model H:} Added BanglaBERT Large Softmax Outputs. Removed K-means of BanglaBERTLarge Embeddings of 23rd layer.\\
\textbf{CRF Model I:} Added All BanglaBERT 1024 Layers Information.

\begin{figure}[h]
\centering
\includegraphics[width=0.5\textwidth]{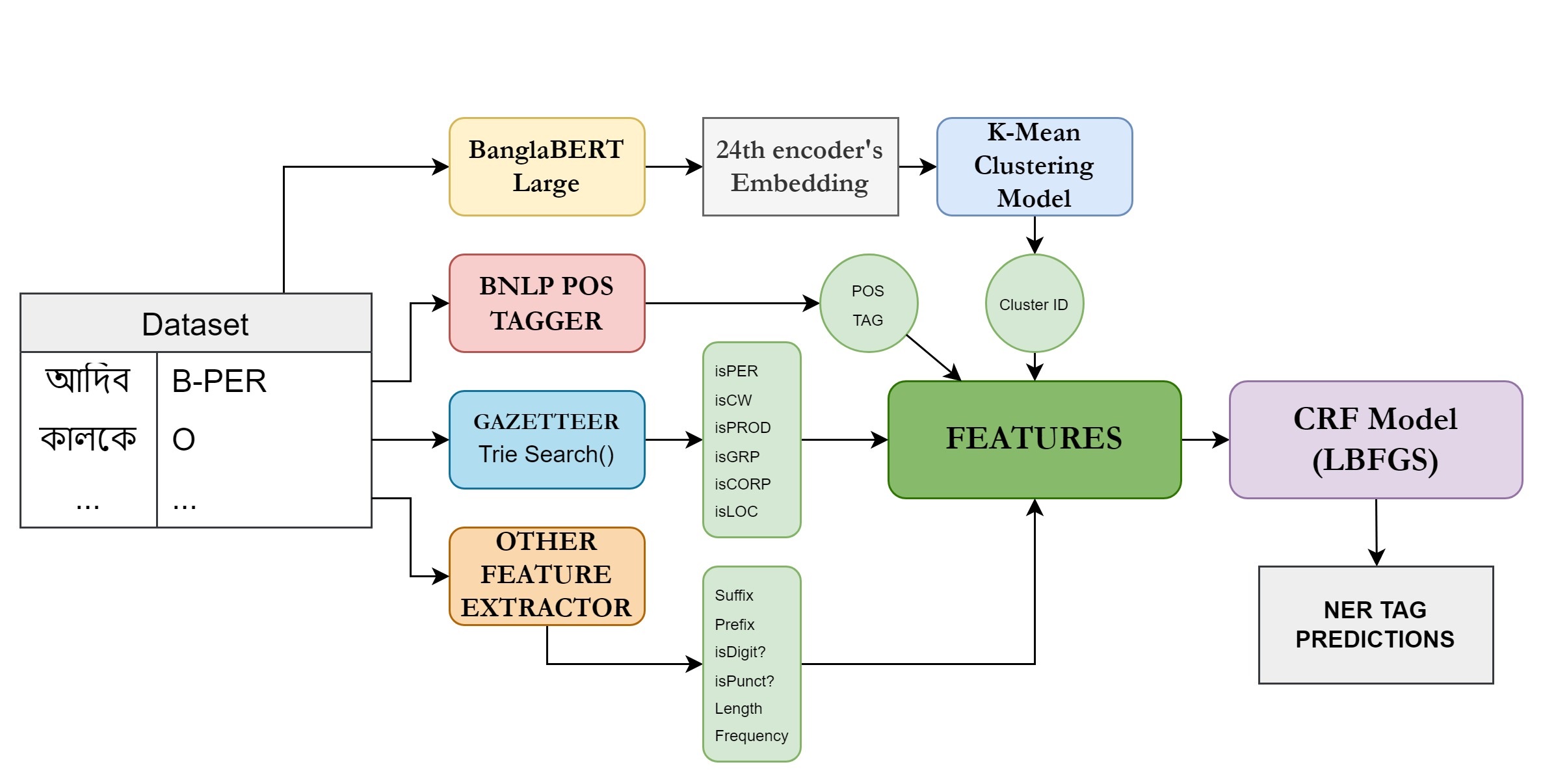}
\caption{CRF Model F}
\label{fig:crfmodelfimage}
\end{figure}

\section{Experiment and Results}
In table \ref{tab:crfmodelsper}, the performance of each strategies of CRF is illustrated.

\begin{table}[h]
\centering
\begin{tabular}{lcc}
\hline
\textbf{Model Name} & \textbf{$F_1$ (Dev)} & \textbf{$F_1$ (Test)} \\
\hline
CRF Model A & 0.671 & 0.3369 \\
CRF Model B & 0.6705 & 0.4379 \\
CRF Model C & 0.6829 & 0.4534 \\
CRF Model D & 0.8103 & 0.8155 \\
CRF Model E & 0.8221 & 0.8074 \\
\textbf{CRF Model F} & 0.815 & \textbf{0.8267} \\
CRF Model G & 0.8172 & 0.8216 \\
\textbf{CRF Model H} & \textbf{0.8467} & 0.8038 \\
CRF Model I & 0.8416 & 0.7398 \\
\hline
\end{tabular}
\caption{CRF Model Strategies Performance in Macro $F_{1}$}
\label{tab:crfmodelsper}
\end{table}

At first, we started with our simple CRF model with basic features in CRF model A. Then, adding more features in CRF model B and CRF model C increases the performance slightly. However, a massive jump is found in CRF model D where we added Gazetteer. Then, in CRF model E, we added isBangla and isStopword features but this decreases our performance. For this reason, we didn't use these features in rest of the strategies. The best performing model in test set was by CRF model F with 0.8267 $F_1$ score. This model used K-means of BanglaBERTLarge Embeddings of 24th Layer. In dev set, CRF model H performed best with 0.8467 $F_1$ score. It used BanglaBERT Large Softmax Outputs.

Now, The table \ref{tab:best} is summarizing the whole results.

\begin{table}[h]
\centering
\begin{tabular}{l c}
\hline
\textbf{Model Name} & \textbf{$F_1$ (Test)} \\
\hline
Two-Layer BiLSTM Network & 0.30 \\
Single Transformer-Based Network & 0.44 \\
BanglaBERT Base (default weight) & 0.2710 \\
BanglaBERT Base (custom weight) & 0.4858 \\
BanglaBERT Large (default weight) & 0.61 \\
BanglaBERT Large (custom weight) & 0.5883 \\
\textbf{CRF Model F} & \textbf{0.8267} \\
\hline
\end{tabular}
\caption{Macro $F_1$ from all of our experiments}
\label{tab:best}
\end{table}

Our baseline models, Two-Layer BiLSTM Network and Single Transformer-Based Network performed 0.3 and 0.44 Macro $F_1$ score in test set. This told us that using only deep learning network would not be enough in BanglaNER task. As transformer is better at capturing context than BiLSTM, the result is slightly higher for transformer model. Additionaly, the introduction of custom weights significantly improved BanglaBERT base performance. However, it slightly reduced BanglaBERT large model's score. So, further experiment is needed for our custom weighted categorical cross entropy loss function which we left as future works. Among the BanglaBERT models, BanglaBERT large performed the best here with 0.61 $F_1$ score. However, beating all the previous models, our CRF model F performed the best with 0.8267 Macro $F_1$ score in MultiCoNER 1 test set. This shows the impact of our knowledge based solution which is Gazetteer in the NER task in Bangla.
\section{Conclusion}
\label{sec:conclusion}
In this research work, we explored the current trends of Bangla Named Entity Recognition tasks. As Bangla is a low-resource language, limited work was done. We had chosen MultiCoNER’s Bangla dataset which is considered as one of the toughest. We conducted experiments with available models and tried to improve them. We proposed a custom loss function to improve BanglaBERT base model performance. Furthermore, we created a Gazetteer containing over 96 thousand entities. The implementation of the Gazetteer showed drastic improvement. Finally, we proposed a complex CRF model which uses the embeddings of the final layer in BanglaBERT with Gazetteer and other features which show promising state-of-the-art performance in Bangla NER tasks.


\bibliography{custom}
\newpage
\section{Appendix}
\label{sec:appendix}

\begin{table*}[h]
\centering
\begin{tabular}{|c|c|}
\hline
\textbf{Name} & \textbf{Features (Added feature are in bold)}                                                                                                                                                                                \\ \hline
CRF Model A   & Suffix, Prefix, Index, Length                                                                                                                                                                                                \\ \hline
CRF Model B   & \begin{tabular}[c]{@{}c@{}}Suffix, Prefix, Index, Length, \textbf{isDigit}, \textbf{isPunctuation}, \\ \textbf{Frequency}\end{tabular}                                                                                                                  \\ \hline
CRF Model C   & \begin{tabular}[c]{@{}c@{}}Suffix, Prefix, Index, Length,  isDigit,  isPunctuation, \\ Frequency, \textbf{POS}\end{tabular}                                                                                                           \\ \hline
CRF Model D   & \begin{tabular}[c]{@{}c@{}}Suffix, Prefix, Index, Length, isDigit, isPunctuation, \\ Frequency, POS, \textbf{Gazetteer}\end{tabular}                                                                                                  \\ \hline
CRF Model E   & \begin{tabular}[c]{@{}c@{}}Suffix, Prefix, Index, Length, isDigit, isPunctuation, \\ \textbf{isBangla}, \textbf{isStopword}, Frequency, POS, \\ Gazetteer\end{tabular}                                                                \\ \hline
CRF Model F   & \begin{tabular}[c]{@{}c@{}}Suffix, Prefix, Index, Length, isDigit, isPunctuation, \\ Frequency, POS, Gazetteer, \textbf{K-means of} \\ \textbf{BanglaBERTLarge Embeddings of 24th Layer}\end{tabular}                                          \\ \hline
CRF Model G   & \begin{tabular}[c]{@{}c@{}}Suffix, Prefix, Index, Length, isDigit, isPunctuation, \\ Frequency, POS, Gazetteer,\textbf{ K-means of} \\ \textbf{BanglaBERTLarge Embeddings} \\ \textbf{of 23rd and 24th Layer}\end{tabular}                              \\ \hline
CRF Model H   & \begin{tabular}[c]{@{}c@{}}Suffix, Prefix, Index, Length, isDigit,  isPunctuation, \\ Frequency, POS, Gazetteer, K-means of \\ BanglaBERTLarge Embeddings of 24th Layer, \\ \textbf{BanglaBERT Large Softmax Outputs}\end{tabular}    \\ \hline
CRF Model I   & \begin{tabular}[c]{@{}c@{}}Suffix, Prefix, Index, Length, isDigit,  isPunctuation, \\ Frequency, POS, Gazetteer, K-means of \\ BanglaBERTLarge Embeddings 24th Layer, \\ \textbf{All BanglaBERT 1024 Layers Information}\end{tabular} \\ \hline
\end{tabular}
\caption{CRF Model Strategies}
\label{fig:crftrats}
\end{table*}

\begin{table*}[h]
\centering
\scalebox{0.84}
{\begin{tabular}{|c|c|c|c|c|}
\hline
\textbf{Model Name}  & \textbf{\begin{tabular}[c]{@{}c@{}}Added Features \\ in the Current Model\end{tabular}}                                & \textbf{\begin{tabular}[c]{@{}c@{}}Removed Features from \\ Previous Model\end{tabular}}                        & \textbf{\begin{tabular}[c]{@{}c@{}}Macro $F_1$ \\ (Dev)\end{tabular}} & \textbf{\begin{tabular}[c]{@{}c@{}}Macro $F_1$ \\ (Test)\end{tabular}} \\ \hline
CRF Model A          & Suffix, Prefix, Index, Length                                                                     &                                                                                             & 0.671                                                              & 0.3369                                                              \\ \hline
CRF Model B          & isDigit, isPunctuation, Frequency                                                                 &                                                                                             & 0.6705                                                             & 0.4379                                                              \\ \hline
CRF Model C          & POS Tag                                                                                           &                                                                                             & 0.6829                                                             & 0.4534                                                              \\ \hline
CRF Model D          & Gazetteer                                                                                         &                                                                                             & 0.8103                                                             & 0.8155                                                              \\ \hline
CRF Model E          & isBangla, isStopword                                                                              &                                                                                             & 0.8221                                                             & 0.8074                                                              \\ \hline
\textbf{CRF Model F} & \begin{tabular}[c]{@{}c@{}}K-Means of BanglaBERT\\ 24th layer Embedding\end{tabular}              & \begin{tabular}[c]{@{}c@{}}isBangla,\\ isStopword\end{tabular}                              & 0.815                                                              & \textbf{0.8267}                                                     \\ \hline
CRF Model G          & \begin{tabular}[c]{@{}c@{}}K-Means of BanglaBERT \\ 23rd and 24th layer \\ Embedding\end{tabular} &                                                                                             & 0.8172                                                             & 0.8216                                                              \\ \hline
CRF Model H          & \begin{tabular}[c]{@{}c@{}}BanglaBERT Large \\ Softmax Output\end{tabular}                        & \begin{tabular}[c]{@{}c@{}}K-Means of \\ BanglaBERT\\ 23rd layer \\ embeddings\end{tabular} & \textbf{0.8467}                                                    & 0.8038                                                              \\ \hline
CRF Model I          & \begin{tabular}[c]{@{}c@{}}BanglaBERT Large 24th \\ layers 1024 sized \\ embedding\end{tabular}   &                                                                                             & 0.8416                                                             & 0.7398                                                              \\ \hline
\end{tabular}}
\caption{CRF Performances}
\label{tab:crfper}
\end{table*}

\end{document}